%% file: unsup_oneshot_2021.tex
\documentclass[a4paper]{article}
\usepackage{INTERSPEECH2021}
\title{Direct multimodal few-shot learning of speech and images}
\name{Leanne Nortje \qquad Herman Kamper}
\address{Department of E\&E Engineering, Stellenbosch University, South Africa}
\email{nortjeleanne@gmail.com, kamperh@sun.ac.za}

\usepackage{graphicx}

\usepackage{booktabs}
\usepackage{tabularx}
\usepackage{multirow}
\newcommand{\mytable}{
    \centering
    \renewcommand{\arraystretch}{1.1}
}

\newcolumntype{C}{>{\centering\arraybackslash}X}
\newcolumntype{L}{>{\raggedright\arraybackslash}X}
\newcolumntype{R}{>{\raggedleft\arraybackslash}X}
\newcolumntype{P}[1]{>{\raggedright\arraybackslash}p{#1}}

\newcommand{\PreserveBackslash}[1]{\let\temp=\\#1\let\\=\temp}
\newcolumntype{A}[1]{>{\PreserveBackslash\raggedright}p{#1}}
\newcolumntype{B}[1]{>{\PreserveBackslash\centering}p{#1}}

\usepackage{amsmath}
\usepackage{amssymb}
\renewcommand{\vec}[1]{\boldsymbol{\mathbf{#1}}}

\usepackage{calc}
\usepackage{microtype}
\usepackage{url}
\usepackage{xcolor}

\makeatletter
\usepackage{acro}
\input{defined_acronyms}

\usepackage[prependcaption,textsize=scriptsize]{todonotes}
\definecolor{hermancolor}{HTML}{008000}
\definecolor{leannecolor}{HTML}{6699CC}
\definecolor{othercolor}{HTML}{CC0000}
\definecolor{oraclecolor}{HTML}{8B8589}

\usepackage{cite}

\let\oldbibliography\thebibliography
\renewcommand{\thebibliography}[1]{\oldbibliography{#1}
    \setlength{\itemsep}{0.4mm}
    \vspace*{-0mm}}

\begin{document}

    \maketitle
    
    \begin{abstract}
        We propose direct multimodal few-shot models that learn a shared embedding space of spoken words and images from only a few paired examples. Imagine an agent is shown an image along with a spoken word describing the object in the picture, e.g.\ \textit{pen}, \textit{book} and \textit{eraser}. After observing a few paired examples of each class, the model is asked to identify the ``book'' in a set of unseen pictures. Previous work used a two-step indirect approach relying on speech-speech and image-image comparisons across the support set of given speech-image pairs. Instead, we propose two direct models which learn a single multimodal space where inputs from different modalities are directly comparable: a \ac{MTriplet} and a \ac{MCAE}. To train these direct models, we \textit{mine} speech-image pairs by using the support set to pair up unlabelled in-domain speech and images. In a speech-to-image digit matching task, direct models outperform indirect models, with the \ac{MTriplet} achieving the best multimodal five-shot accuracy. We show that the improvements are due to the combination of unsupervised and transfer learning in the direct models, and the absence of two-step compounding errors.
    \end{abstract}
    
    %% SPL
    %\begin{IEEEkeywords}
    %	few-shot learning, multimodal modelling, unsupervised learning, transfer learning, speech and images
    %\end{IEEEkeywords}
    %\IEEEpeerreviewmaketitle
    
    % Interspeech
    \noindent\textbf{Index Terms}: few-shot learning, multimodal modelling, unsupervised learning, transfer learning, speech and images.

    \acresetall
    \input{Sections/Introduction}
    \input{Sections/Multimodal_task}
    \input{Sections/Unimodal_features}
    \input{Sections/Multimodal_features}

    \input{Sections/Experimental_setup}
    \input{Sections/Experimental_results}
    \input{Sections/Conclusion}

    %\newpage
    \bibliography{library}
    
\end{document}

%% file: defined_acronyms.tex
\DeclareAcronym{AE}{
	short = AE ,
	long  = autoencoder
}
\DeclareAcronym{CAE}{
	short = CAE,
	long  = correspondence autoencoder,
	alt = Correspondence Autoencoder
}
\DeclareAcronym{MCAE}{
	short = MCAE,
	long  = multimodal correspondence autoencoder,
	alt = Multimodal Correspondence Autoencoder
}
\DeclareAcronym{MTriplet}{
	short = MTriplet,
	long  = multimodal triplet network,
	alt = Multimodal Triplet Network
}
\DeclareAcronym{FFNN}{
	short = FFNN,
	long  = feedforward neural network,
	alt = Feedforward Neural Networks
}
\DeclareAcronym{CNN}{
	short = CNN,
	long  = convolutional neural network,
	alt = Convolutional Neural Networks
}
\DeclareAcronym{RNN}{
	short = RNN,
	long  = recurrent neural network,
	alt = Recurrent Neural Networks
}
\DeclareAcronym{MFCC}{
	short = MFCC,
	long  = mel-frequency cepstral coefficient
}
\DeclareAcronym{GRU}{
	short = GRU,
	long  = gated recurrent unit
}
\DeclareAcronym{DTW}{
	short = DTW,
	long  = dynamic time warping,
	alt = Dynamic Time Warping
}
\DeclareAcronym{ML}{
	short = ML,
	long  = machine learning
}
\DeclareAcronym{DL}{
	short = DL,
	long  = deep learning
}
\DeclareAcronym{HBPL}{
	short = HBPL,
	long  = hierarchical Bayesian program learning
}
\DeclareAcronym{HHMM}{
	short = HHMM,
	long  = hierarchichal hidden Markov model
}
\DeclareAcronym{LSTM}{
	short = LSTM, 
	long = long short-term memory
}
\DeclareAcronym{SGD}{
	short = SGD, 
	long = stochastic gradient decent,
	alt = Stochastic Gradient Decent
}
\DeclareAcronym{MAML}{
	short = MAML, 
	long = model-agnostic meta-learning
}
\DeclareAcronym{VAE}{
	short = VAE, 
	long = variational autoencoder
}

%% file: Sections/Introduction.tex
\section{Introduction}
\label{sec:intro}

Current audio and vision systems require large amounts of labelled data which is expensive and time-consuming to collect.
In contrast, young children are able to learn new words and objects from only a few examples~\cite{biederman_recognition-by-components:_1987, miller_how_1987,gomez_infant_2000,lake+etal_cogsci14,rasanen_joint_2015}.
In fact, from only one exposure, infants can learn the word for a shown (visual) object~\cite{borovsky_once_2012}.
Can we emulate this in a machine?
Imagine an agent is shown images of a \textit{duck}, \textit{horse} and \textit{chicken}, where each image is paired with a spoken word describing the object.
After observing a few example pairs from each class, the agent is prompted to identify the visual instance corresponding to the spoken word ``horse'' from a set of unseen images. 
This task was formalised in~\cite{eloff_multimodal_2019}: \textit{multimodal few-shot learning} is the task of learning new concepts from a few paired examples, where each pair consists of two items of the same concept but in different modalities.

Previous work~\cite{eloff_multimodal_2019,nortje_unsupervised_2020} used a two-step \textit{indirect} approach: % where two separate unimodal \herman{representation spaces} %networks are used:
a spoken query is compared to the spoken examples in the given \textit{support set} of speech-image pairs, and the corresponding image is then used to select the closest item in the unseen \textit{matching set}.
The task is therefore reduced to two unimodal comparisons, with the support set acting as a pivot between the modalities. 
To do the unimodal comparisons, learned speech and image representations are used:
%For these unimodal comparisons,
\cite{nortje_unsupervised_2020} compared \textit{transfer learning} to \textit{unsupervised learning}.
In transfer learning, representations are used from models trained on labelled background data (not containing any classes seen at test time).
The motivation for this
is that humans can call on prior knowledge when
learning new concepts.
In unsupervised modelling, models are trained on unlabelled in-domain data.
This is reasonable since,
before being shown paired examples, an agent could be exposed to a large amount of unlabelled
speech and visual data from its environment. 

In this paper we propose combining unsupervised and transfer learning to obtain \textit{direct} models that learn a single multimodal embedding space where observations from either modality can be directly compared, i.e.\  word and image instances of the same class are mapped to similar representations in a shared space.
Training such a multimodal speech-image space has been considered in several studies~\cite{harwath_unsupervised_2016,harwath_jointly_2018,harwath_learning_2020,leidal_learning_2017,chrupala+etal_acl17,scharenborg+etal_icassp18,merkx+etal_arxiv19,havard+etal_arxiv20}. But they all train models on a large set of paired speech-image examples for a large number of classes. We instead focus on how a multimodal space can be learned from only a few examples for a limited set of classes.

To train direct models, we require matched speech-image pairs.
Since the given 
pairs in the support set is not sufficient to train models directly, we propose to \textit{mine} speech-image training pairs. 
Spoken/visual items in the support set are compared to unlabelled examples in the respective modalities. New training speech-image pairs are then constructed by using the support set as a pivot.
We are therefore making use of unlabelled in-domain data (a form of unsupervised learning).
But to do the speech-speech and image-image comparisons in the mining scheme, we actually use representations from transfer-learned speech or vision classifiers (trained on background labelled data).
Direct modelling thus combines unsupervised and transfer learning.

We use the mined cross-modal pairs to train two direct
models: a \ac{MTriplet}, which combines two unimodal triplet losses, and a novel \ac{MCAE}, which combines two unimodal \acp{CAE}.
A CAE attempts to produce another 
instance of the same class as the input 
through a bottleneck 
layer~\cite{kamper_unsupervised_2015}.
The \ac{MCAE} is similar to other multimodal models, although previous work mostly used \acp{AE} and worked in other modalities~\cite{weston_large_2010,ngiam_multimodal_2011,frome_devise_2013,socher_zero-shot_2013,silberer_learning_2014,feng_cross-modal_2014}.
%The \ac{MTriplet} is similar to the models of~\cite{harwath_unsupervised_2016,leidal_learning_2017,harwath_jointly_2018,harwath_learning_2020}.
%But we are the first to use these model architectures in a multimodal few-shot learning setting.
The \ac{MTriplet} is similar to the models of~\cite{harwath_unsupervised_2016,leidal_learning_2017,harwath_jointly_2018,harwath_learning_2020}, but we are first to use these architectures in a multimodal few-shot learning setting.
\cite{pahde_multimodal_2021, huang_acmm_2019} used images with text captions on a similar multimodal few-shot task.

In a speech-to-image digit matching task, we compare our direct models to the best indirect transfer-learned and unsupervised models from~\cite{nortje_unsupervised_2020}.
Both direct models outperform the indirect models, with the \ac{MTriplet} as the top performer.
Ablation experiments show that the direct models' superior performance can be attributed to the way in which they combine unsupervised and transfer learning, and because direct matching does not cause a compounding of errors as in the two-step indirect methods.\footnote{\raggedright We release source code at: {\scriptsize \url{https://github.com/LeanneNortje/direct_multimodal_few-shot_learning}}.}

%% file: Sections/Multimodal_task.tex
\begin{figure*}[!t]
	
	%	\begin{minipage}[b]{1.0\linewidth}
	\centering
	\includegraphics[width=0.95\linewidth]{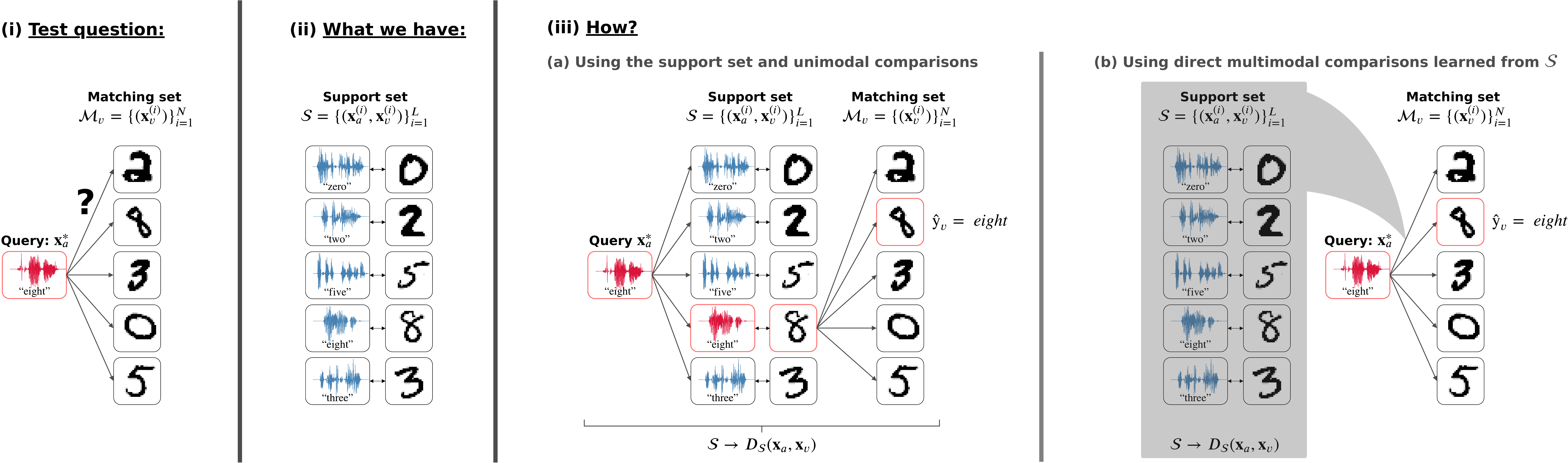}
	\vspace*{-7.5pt}
	%		\vspace*{-20pt}
	%	\end{minipage}
	
	\caption{(i) The multimodal one-shot speech-to-image matching question shown at test time.
		(ii) The support set given to solve the task. (iii) The support set can be used in two ways: (a) an indirect matching approach~\cite{eloff_multimodal_2019,nortje_unsupervised_2020} which uses two unimodal comparisons across the support set, and (b) the direct matching approach proposed in this paper.}
	\label{fig:few_shot_task}
	
	\vspace*{-7pt}
	
\end{figure*}

%\section{The Task: Multimodal Few-Shot Speech-to-Image Matching}\label{sec:multimodal_task}
\section{Task: Multimodal few-shot speech-to-image matching}\label{sec:multimodal_task} % Interspeech

The task of multimodal few-shot matching of speech and images is illustrated in Figure~\ref{fig:few_shot_task}\,(i). We are given an unseen speech query $\mathbf{x}^*_{a}$ and prompted to identify the corresponding image in a matching set $\mathcal{M}_v=\{(\boldsymbol{\mathrm{x}}_v^{(i)})\}_{i=1}^N$ of unseen test images.
To do this,
we are given a small number of pairs in
a \textit{support set} $\mathcal{S}$, where each pair consists of an isolated spoken word $\boldsymbol{\mathrm{x}}_a^{(i)}$ and a corresponding image $\boldsymbol{\mathrm{x}}_v^{(i)}$.
Neither the test-time speech query $\mathbf{x}^*_{a}$ nor the matching set $\mathcal{M}_v$ images occur in the support set.
For the \textit{one-shot} case shown in Figure~\ref{fig:few_shot_task}\,(ii), $\mathcal{S}$ consists of one example pair for each of the $L$ classes. 
Multimodal one-shot matching can be extended to \textit{$K$-shot} matching:
for $L$-way $K$-shot matching, the support set $\mathcal{S}=\{(\boldsymbol{\mathrm{x}}_a^{(i)}, \boldsymbol{\mathrm{x}}_v^{(i)})\}_{i=1}^{L \times K}$ contains $K$ paired speech-image examples for each of the $L$ classes. 

To perform this matching task, we need a distance metric $D_\mathcal{S}(\boldsymbol{\mathrm{x}}_a, \boldsymbol{\mathrm{x}}_v)$ between inputs from the two modalities.
Previous work, which we use as a baseline (\S\ref{sec:unimodal_features}), constructed an indirect distance metric while we propose a direct method (\S\ref{sec:multimodal_features}).

%% file: Sections/Unimodal_features.tex
%\section{Baseline: Indirect Multimodal Few-shot Matching Approach}\label{sec:unimodal_features}
\section{Baseline: Indirect multimodal %few-shot
matching}\label{sec:unimodal_features} % Interspeech

Indirect %multimodal few-shot
matching uses two unimodal comparisons across the support set as its metric $D_\mathcal{S}(\boldsymbol{\mathrm{x}}_a, \boldsymbol{\mathrm{x}}_v)$, as shown in Figure~\ref{fig:few_shot_task}\,(iii)\,(a).
A spoken query $\mathbf{x}^*_{a}$ is encoded into some representation and then compared to representations for each speech instance $\boldsymbol{\mathrm{x}}_a^{(i)}$ in $\mathcal{S}$ to determine the query's closest spoken match based on the (cosine) distance between representations.
The nearest neighbour's paired image is then compared to each image instance $\boldsymbol{\mathrm{x}}_v^{(i)}$ in the matching set $\mathcal{M}_v$ to find its closest image based on the distance between image representations.
The closest image is the model's
prediction, e.g. in Figure~\ref{fig:few_shot_task}\,(iii)\,(a) %it is
the %image of the
\textit{eight} in $\mathcal{M}_v$. 

This indirect approach requires unimodal representations that capture similarity within a modality. 
In~\cite{nortje_unsupervised_2020}, two methodologies were considered. % for unimodal comparisons.
The first is \textit{transfer learning}, which is common in unimodal few-shot learning studies~\cite{li_fei-fei_bayesian_2003,fei-fei_one-shot_2006,lake_one-shot_2013,lake_human-level_2015,koch_siamese_2015,vinyals_matching_2016,parnami_few-shot_2020}.
%Transfer learning is a method of
This involves
training a model on a different but related dataset not containing any of the classes seen at test time~\cite{pan_survey_2009,ruder_neural_2019}.
\cite{nortje_unsupervised_2020} specifically
considered unimodal transfer-learned speech and vision classifiers, Siamese~\cite{bromley_signature_1994} triplet networks, and \acp{CAE}. All these models were trained on labelled background data not containing any instances of the few-shot classes seen at test time. 

The second unimodal representation learning approach is \textit{unsupervised learning}. Before an agent is shown paired examples of new classes, the agent could be exposed to 
unlabelled speech and visual data from its environment.
Some of these unlabelled examples could correspond to the few-shot classes. % seen at test time.
Therefore, \cite{nortje_unsupervised_2020} also considered %also train separate
unsupervised unimodal speech and vision \acp{AE} and \acp{CAE} trained on unlabelled in-domain data.

%% file: Sections/Multimodal_features.tex
%\section{Direct Multimodal Matching using Features from a Multimodal Space}\label{sec:multimodal_features}
\section{Proposal: Direct multimodal matching}\label{sec:multimodal_features} % Interspeech

\begin{figure*}[!t]
	
%	\begin{minipage}[b]{1.0\linewidth}
		\centering
		\includegraphics[width=0.8\linewidth]{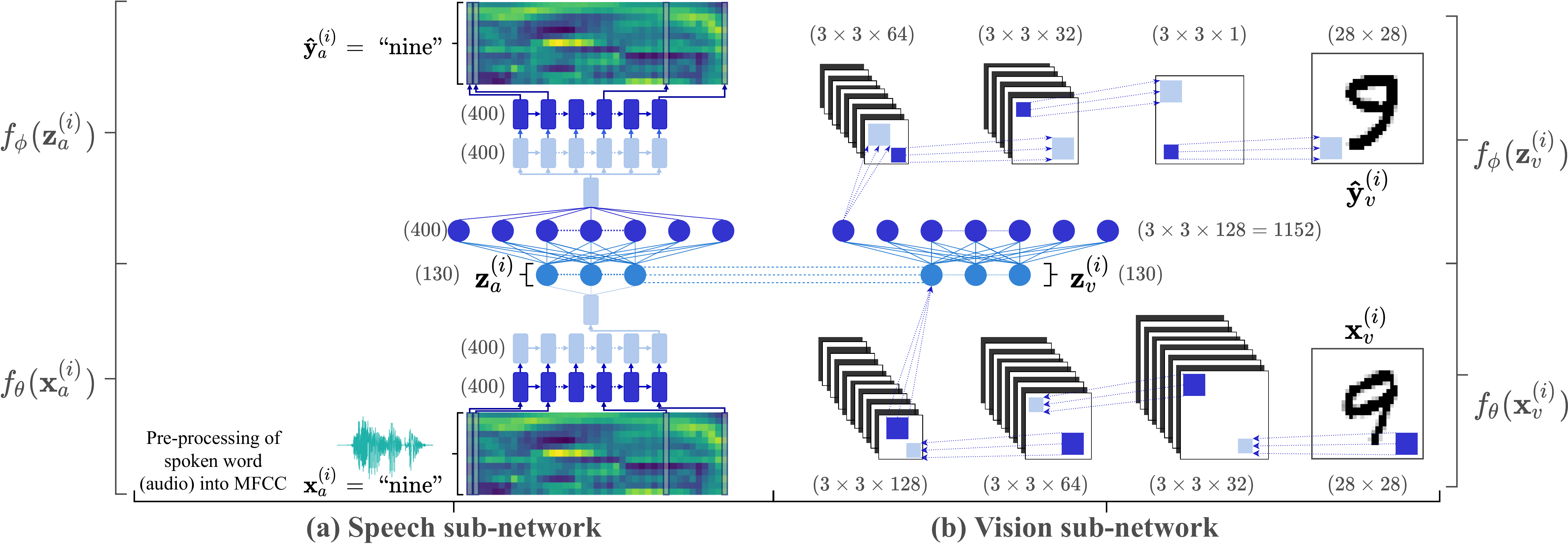}
		\vspace*{-7.5pt}
%		\vspace*{-10pt}
%	\end{minipage}
	
	\caption{The \ac{MCAE} uses (a) an \ac{RNN} encoder-decoder for processing spoken words, and (b) a \ac{CNN} encoder-decoder for processing images. The  \ac{MCAE} loss is a combination of \ac{CAE} losses in each modality and a multimodal loss encouraging the audio and visual latent representations $\mathbf{z}_a^{(i)}$ and $\mathbf{z}_v^{(i)}$ to be close to each other.}
	\label{fig:arch}
	
	\vspace*{-12pt}
	
\end{figure*}

A direct multimodal few-shot learning model aims to learn a single multimodal embedding space which maps inputs of the same class, regardless of modality, to similar representations.
As illustrated in Figure~\ref{fig:few_shot_task}\,(iii)\,(b), the multimodal space can be used as a direct cross-modal metric $D_\mathcal{S}(\boldsymbol{\mathrm{x}}_a, \boldsymbol{\mathrm{x}}_v)$, allowing a speech query's representation to be directly compared to the image representations in the matching set.

\subsection{Cross-modal pair mining}
\label{sec:mining}

In order to learn a multimodal embedding space, % train a direct multimodal few-shot learning model,
we need pairs of matching images and spoken words.
The only pairs we have are those in the given support set $\mathcal{S}$.
This small set is not sufficient for training a multimodal model.
We therefore use $\mathcal{S}$ to \textit{mine} more in-domain speech-image pairs from a larger set of unlabelled data, using $\mathcal{S}$ as a pivot between unlabelled data in the two modalities.
This is similar to ideas used in~\cite{cong+etal_arxiv20,bhosale+etal_arxiv21}.
Consider the $i^\textrm{th}$ speech-image pair in the support set $(\boldsymbol{\mathrm{x}}_a^{(i)}, \boldsymbol{\mathrm{x}}_v^{(i)})$.
We find the images in the unlabelled in-domain image data whose closest image in the support set is $\boldsymbol{\mathrm{x}}_v^{(i)}$.
Similarly, for spoken words in unlabelled in-domain speech data, we find those for which $\boldsymbol{\mathrm{x}}_a^{(i)}$ is the closest match.
From these items %unlabelled  word and image instances
matched to the $i^\textrm{th}$ pair, we choose a random word and image instance and pair them up, thereby obtaining a novel training pair.
Since we use no labels, all the pairs will not be correct; nevertheless, this procedure provides a large number of noisy training examples.

Similar to the indirect matching approach (\S\ref{sec:unimodal_features}), we need to do unimodal speech-speech and image-image comparisons to perform mining across $\mathcal{S}$.
We employ transfer learning, specifically the best unimodal models from \cite{nortje_unsupervised_2020}: the transfer-learned speech and vision classifiers.
For the support set and all the unlabelled speech and image items, we therefore extract features from an intermediate embedding (the layer before the softmax) and then use cosine distance to find the closest matches. % to represent the given input.
Mining thus combines unsupervised learning (from the unlabelled in-domain data) and transfer learning (from background out-of-domain data).
Our experiments (\S\ref{sec:experiments}) show that using unimodal transfer-learned models is essential to the mining process.

We consider two multimodal architectures trained on these mined speech-image pairs: a novel \ac{MCAE} and an \ac{MTriplet}.

\subsection{\Acl{MCAE}}
\label{sec:mcae}

The \acf{MCAE} aims to learn a multimodal space by using a modified \acf{CAE} loss.
A standard \ac{CAE} is trained to produce another instance $\mathbf{x}_{\textrm{pair}}^{(i)}$ of the same type as the input $\mathbf{x}^{(i)}$ through a bottleneck layer~\cite{kamper_unsupervised_2015}.
The \ac{CAE} loss is $\ell = ||\mathbf{x}_{{\textrm{pair}}}^{(i)} - \hat{\boldsymbol{\mathrm{y}}}^{(i)}||^2_2$, where $\hat{\boldsymbol{\mathrm{y}}}^{(i)}$ is the network
output. 

The \ac{MCAE} is a combination of a speech \ac{CAE} and a vision \ac{CAE}, as shown in Figure~\ref{fig:arch}. Each \ac{CAE} consists of an encoder and a decoder.
The \acf{RNN} speech encoder $f_\theta(\boldsymbol{\mathrm{x}}_a^{(i)})$ encodes input $\mathbf{x}_a^{(i)}$ to the representation $\boldsymbol{\mathrm{z}}_a^{(i)}$.
The \ac{RNN} decoder $f_\phi(\boldsymbol{\mathrm{z}}_a^{(i)})$ is conditioned on $\boldsymbol{\mathrm{z}}_a^{(i)}$ to produce output $\hat{\boldsymbol{\mathrm{y}}}_a^{(i)}$. 
Similar to the acoustic embedding models of~\cite{chung_unsupervised_2016,wang_segmental_2018,holzenberger_learning_2018,kamper_truly_2019,peng+etal_sas20}, the encoder produces a fixed-sized representation $\boldsymbol{\mathrm{z}}_a^{(i)}$ for variable duration input $\boldsymbol{\mathrm{x}}_a^{(i)}$.
The speech CAE's loss is $\ell_a = ||\mathbf{x}_{a_{\textrm{pair}}}^{(i)} - \hat{\boldsymbol{\mathrm{y}}}_a^{(i)}||^2_2$.  
The vision \ac{CAE} has a similar structure and loss $\ell_v$, but uses a \acf{CNN} encoder and a transposed convolutional vision decoder instead of \acp{RNN}.

The \ac{MCAE} links %by linking
the speech and vision \ac{CAE}'s with a multimodal loss term
that encourages similar latent representations for paired inputs:
$\ell_z = ||\boldsymbol{\mathrm{z}}_a^{(i)} - \boldsymbol{\mathrm{z}}_v^{(i)}||^2_2$.
The complete MCAE loss for a singular training example is $\ell = \alpha_a\ell_a + \alpha_v\ell_v + \alpha_z\ell_z$, where the $\alpha$'s are loss weights. 
%Therefore, 
Each training example therefore consists of $\mathbf{x}_a^{(i)}$, $\mathbf{x}_{a_{\textrm{pair}}}^{(i)}$, $\mathbf{x}_v^{(i)}$ and $\mathbf{x}_{v_{\textrm{pair}}}^{(i)}$, where $(\mathbf{x}_a^{(i)}, \mathbf{x}_v^{(i)})$ is a mined speech-image pair.
For $\mathbf{x}_a^{(i)}$ we need the paired word $\mathbf{x}_{a_{\textrm{pair}}}^{(i)}$, and for $\mathbf{x}_v^{(i)}$ we need the paired image $\mathbf{x}_{v_{\textrm{pair}}}^{(i)}$.
Again we do not have class labels for the unlabelled in-domain data, so we mine these unimodal pairs. The procedure is similar to that of the cross-modal mining scheme in \S\ref{sec:mining}, but here we mine pairs within the modality, so the support set is not used as a pivot.
As noted in \S\ref{sec:intro}, the overall \ac{MCAE} model is very similar to earlier \ac{AE}-based models, such as the one from~\cite{silberer_learning_2014}.

\subsection{\Acl{MTriplet}}

The \acf{MTriplet} is based on the model of~\cite{harwath_unsupervised_2016}. But here we apply it for the first time to multimodal few-shot learning.
The \ac{MTriplet} aims to learn a %relative
distance metric between spoken words and images in a single multimodal space by pushing paired cross-modal representations toward each other while pulling non-matching representations away from each other.
The network consists of speech and vision encoders which maps inputs $\mathbf{x}^{(i)}$ to representations $\boldsymbol{\mathrm{z}}^{(i)}$.
In our case, these encoders have exactly the same architecture as the encoders of the \ac{MCAE} shown in Figure~\ref{fig:arch}. %(without decoders).
The model %\ac{MTriplet}
optimises the combination of two triplet losses~\cite{chechik_large_2009,wang_learning_2014,hermann_multilingual_2014,hoffer_deep_2015} so that the distance between representations of paired cross-modal inputs $(\mathbf{x}_a^{(i)}, \mathbf{x}_v^{(i)})$ are smaller than the distance between representations of non-matched cross-modal pairs $(\mathbf{x}_a^{(i)}, \mathbf{x}_{v_{\textrm{neg}}}^{(i)})$ and $(\mathbf{x}_{a_{\textrm{neg}}}^{(i)}, \mathbf{x}_v^{(i)})$, by some margin.
Stated mathematically, the loss is:\vspace*{-5pt}
\begin{align*}
\ell = &\max\left\{0, m + d(\mathbf{z}_a^{(i)}, \mathbf{z}_v^{(i)}) - d(\mathbf{z}_a^{(i)}, \mathbf{z}_{v_{\textrm{neg}}}^{(i)})\right\} + \\[-2pt]
& \max\left\{0, m + d(\mathbf{z}_a^{(i)}, \mathbf{z}_v^{(i)}) - d(\mathbf{z}_{a_{\textrm{neg}}}^{(i)},\mathbf{z}_v^{(i)})\right\}, \\[-19pt]
\end{align*}
where $m$ is the margin and $d(\cdot, \cdot)$ is the cosine distance~\cite{harwath_unsupervised_2016}.
Each \ac{MTriplet} training example consists of items $\mathbf{x}_a^{(i)}$, $\mathbf{x}_{a_{\textrm{neg}}}^{(i)}$, $\mathbf{x}_v^{(i)}$ and $\mathbf{x}_{v_{\textrm{neg}}}^{(i)}$, where $(\mathbf{x}_a^{(i)}, \mathbf{x}_v^{(i)})$ are mined speech-image pairs and $\mathbf{x}_{a_{\textrm{neg}}}^{(i)}$ and $\mathbf{x}_{v_{\textrm{neg}}}^{(i)}$ are negative examples.
We mine negatives similarly to the within-modality positives (end of \S\ref{sec:mcae}), but we add constraints inspired by~\cite{schroff_facenet:_2015, hoffer_deep_2015,hermans_defense_2017,jansen_unsupervised_2018} to obtain ``hard'' negatives, i.e.\ they are non-matching but close to the items in $(\mathbf{x}_a^{(i)}, \mathbf{x}_v^{(i)})$.

%% file: Sections/Experimental_setup.tex
%\section{Experimental Setup}
\section{Experimental setup} % Interspeech

%\subsection{Data}
\textbf{Data.}
As in~\cite{eloff_multimodal_2019,nortje_unsupervised_2020}, %our experiments are based on the
multimodal few-shot tasks are constructed by pairing isolated spoken digits from the TIDigits corpus~\cite{leonard_r._gary_tidigits_1993} with handwritten digit images from MNIST~\cite{lecun_yann_gradient-based_1998}.
TIDigits contains speech from 326 speakers. 
We use the test subsets of TIDigits and MNIST for the multimodal matching task.
For training unsupervised indirect models (\S\ref{sec:unimodal_features}) and for mining pairs in the direct models (\S\ref{sec:mining}), we use the training subsets as unlabelled in-domain data by removing the class labels. For validating these models, we use the validation subsets (again without labels).

For background 
speech data we use the Buckeye corpus of English conversational speech from 40 speakers~\cite{pitt_mark_a._buckeye_2005}.
All utterances (from both TIDigits and Buckeye) are split into isolated words using forced alignments and 
parametrised as static \acp{MFCC}.
For background image data, we use the Omniglot dataset of grayscale images from 1623 handwritten character classes~\cite{lake_human-level_2015}, which we invert and downsample to $28 \times 28$ pixels (to match the MNIST format).
All image pixels are normalised to $[0, 1]$.
Transfer learning is performed using the labelled training subsets of Buckeye and Omniglot, with validation subsets used for validation.
We ensure that the background data does not contain any digit classes. % instances of digits.

%\subsection{Models}
\textbf{Models.}
For the indirect models (\S\ref{sec:unimodal_features}) we use the same architectures as in \cite{nortje_unsupervised_2020}.
For the direct models (\S\ref{sec:multimodal_features}), we mine pairs via a five-shot support set, i.e.\ five word-image pairs for each of the classes.
The \ac{MCAE} architecture is given in Figure~\ref{fig:arch}, with the loss weights set to $\alpha_a = 0.3$, $\alpha_v = 0.3$ and $\alpha_z = 0.4$. 
The \ac{MTriplet} architecture is identical to the speech and vision encoders in Figure~\ref{fig:arch}; we use a margin of $m = 0.2$. 
We use 130-dimensional representations $\mathbf{z}$ throughout to allow for fair comparisons.
Neural networks are implemented in TensorFlow and trained with a learning rate of $10^{-3}$ using Adam optimisation~\cite{kingma_adam_2015}.
The above hyperparameters are based on previous work~\cite{harwath_unsupervised_2016,nortje_unsupervised_2020} and weren't explicitly tuned.
We did consider the effect of batch size; despite robust performance, 
we report model stability over five different sizes (from 16 to 256).
Early stopping is performed by tracking model loss on  mined speech-image validation pairs (so no labels are used for validation either).

%\subsection{Evaluation}
\textbf{Evaluation.}
As in~\cite{nortje_unsupervised_2020}, we evaluate our models on $400$ five-shot \textit{episodes}~\cite{vinyals_matching_2016}.
Each episode's support set contains five word-image pairs for each of the $L = 11$ digit classes (``one" to ``nine", as well as ``zero" and ``oh"), i.e.\ a five-shot 11-way task.
For each episode's matching set, we sample ten digit images not in the support set.
If the speech query is either a ``zero" or an ``oh", it is counted as correct if the model predicts the matching image to be a $0$.
In each episode, we sample ten different spoken digit queries (not in the support set).
The \ac{MCAE} and \ac{MTriplet} scores are averaged over five different batch sizes, each trained with five different seeds (i.e.\ 25 models in total in each case).
Scores are reported with 95\% confidence intervals.

%% file: Sections/Experimental_results.tex
%\section{Experimental Results}
\section{Experimental results} % Interspeech
\label{sec:experiments}

\begin{table}[!b]
	\vspace*{-7pt}
	\mytable
	%	\caption{Multimodal Five-Shot Speech-to-Image Matching.}\\
	\caption{Multimodal five-shot 11-way speech-to-image matching accuracy (\%).
	Direct models are either trained with pairs mined using cosine distance over the input features or with representations from transfer-learned models.
	Oracle results with ground truth pairs are given for reference.
	}
	\eightpt % Interspeech
	\vspace*{-7.5pt}
	\begin{tabularx}{1.0\linewidth}{@{}Llc@{}}
		\toprule
		& Model & Accuracy \\
		%		\multicolumn{2}{c}{Model} & 11-way five-shot\\
		%		 & & accuracy (\%)\\
		\midrule
		\multirow{3}{8em}{\centering Indirect multimodal few-shot models}
		& DTW + pixels~\cite{eloff_multimodal_2019,nortje_unsupervised_2020} & 41.9 \hphantom{$\pm$ 1.7}\\
		& Transfer-learned classifier~\cite{eloff_multimodal_2019,nortje_unsupervised_2020} & 59.7 $\pm$ 1.7\\
		& Unsupervised CAE~\cite{nortje_unsupervised_2020} & 52.2 $\pm$ 0.7\\
		\addlinespace
		\multirow{6}{8em}{\centering Direct multimodal few-shot models}
		& \ac{MCAE} (cosine mined pairs) & 59.1 $\pm$ 3.1 \\
		& \ac{MCAE} (transfer learning pairs) & 74.9 $\pm$ 1.9 \\
		& \textcolor{oraclecolor}{Oracle \ac{MCAE}} & \textcolor{oraclecolor}{93.6 $\pm$ 1.6}\\[2pt]
%		\addlinespace
		& \ac{MTriplet} (cosine mined pairs) & {67.3 $\pm$ 2.0} \\
		& \ac{MTriplet} (transfer learning pairs) & \textbf{85.5 $\pm$ 1.6} \\
		& \textcolor{oraclecolor}{Oracle \ac{MTriplet}} & \textcolor{oraclecolor}{99.1 $\pm$ 0.1} \\
		\bottomrule
	\end{tabularx}
	\label{tbl:multimodal}
	%	\vspace*{-12pt}
\end{table}

Multimodal five-shot matching results are shown in Table~\ref{tbl:multimodal}.
The first row is a naive indirect baseline where matching is performed on the input features: \acf{DTW} over MFCCs for speech and cosine distance over image pixels.
The best overall score of 85.5\% is achieved by the direct \ac{MTriplet}, giving an absolute improvement of more than 25\% over the best previous result~\cite{nortje_unsupervised_2020}.
This model is followed by the direct \ac{MCAE}.

Table~\ref{tbl:multimodal} also shows scores when pair mining (\S\ref{sec:mining}) is performed using cosine distance over the original features rather than using transfer-learned speech and image classifiers.
The latter gives substantially better scores in both the \ac{MCAE} and \ac{MTriplet}.
The combination of unsupervised learning (mining pairs from unlabelled in-domain data) with transfer learning (using a metric learned from labelled background data) is therefore essential for the superior performance of these direct models.
Oracle results (when perfect pairs are used) also show that improvements to the mining approach (potentially by improving transfer learning) could lead to further future improvements.

Both~\cite{eloff_multimodal_2019} and \cite{nortje_unsupervised_2020} showed that many of the errors from the indirect models are due to mistakes in speech-speech unimodal comparisons which compound with mistakes in the subsequent image-image matching step.
The direct models do not suffer from this, since the two-step approach is replaced with a single cross-modal comparison in the multimodal space.
The question then is how much of the direct models' improvements are due to better comparisons within the multimodal space, and how much are due to not having two comparisons?
To quantify this, we use the direct models, but instead of performing direct comparisons, we use them as though they are indirect models, i.e.\ we use the multimodal space separately to do unimodal speech-speech and image-image comparisons via the support set (\S\ref{sec:unimodal_features}).
Table~\ref{tbl:support_set} shows that there is indeed a large gain by not having to do two comparisons (row 1 vs row 2).
But even the indirect application of the direct models (row 2) are better than the indirect models (Table~\ref{tbl:multimodal}), showing that the \ac{MCAE} and \ac{MTriplet} learn better representations even for within-modality comparisons.

\begin{table}[t!]
	\mytable
	%	\caption{Multimodal Speech-to-Image Matching of the Direct Models on the Direct vs.\ Indirect Approach on a One-Shot Support Set.}
	\caption{Multimodal speech-to-image matching accuracy (\%) when direct models are either used directly (as intended) or as though they are indirect models (i.e.\ two separate unimodal comparisons in the multimodal space via the support set).}
	\eightpt % Interspeech
	\vspace*{-7.5pt}
	\begin{tabularx}{1.0\linewidth}{@{}lCC@{}}
		\toprule
		\addlinespace
		\multirow{2}{*}{Application mode} & \multicolumn{2}{c}{Model}\\
		& \ac{MCAE} & \ac{MTriplet} \\
		\midrule
		\addlinespace
		Direct cross-modal comparisons & \textbf{74.9 $\pm$ 1.9} & \textbf{85.5 $\pm$ 1.4} \\
		%		\addlinespace
		Indirect unimodal comparisons & 66.4 $\pm$ 2.6 & 76.2 $\pm$ 1.5 \\
		%		\addlinespace
		\bottomrule
	\end{tabularx}
	\label{tbl:support_set}
	%	\vspace*{-12pt}
    \vspace*{-2pt}
\end{table}

\begin{figure}[!t]
	\centering
	\includegraphics[width=0.7\linewidth]{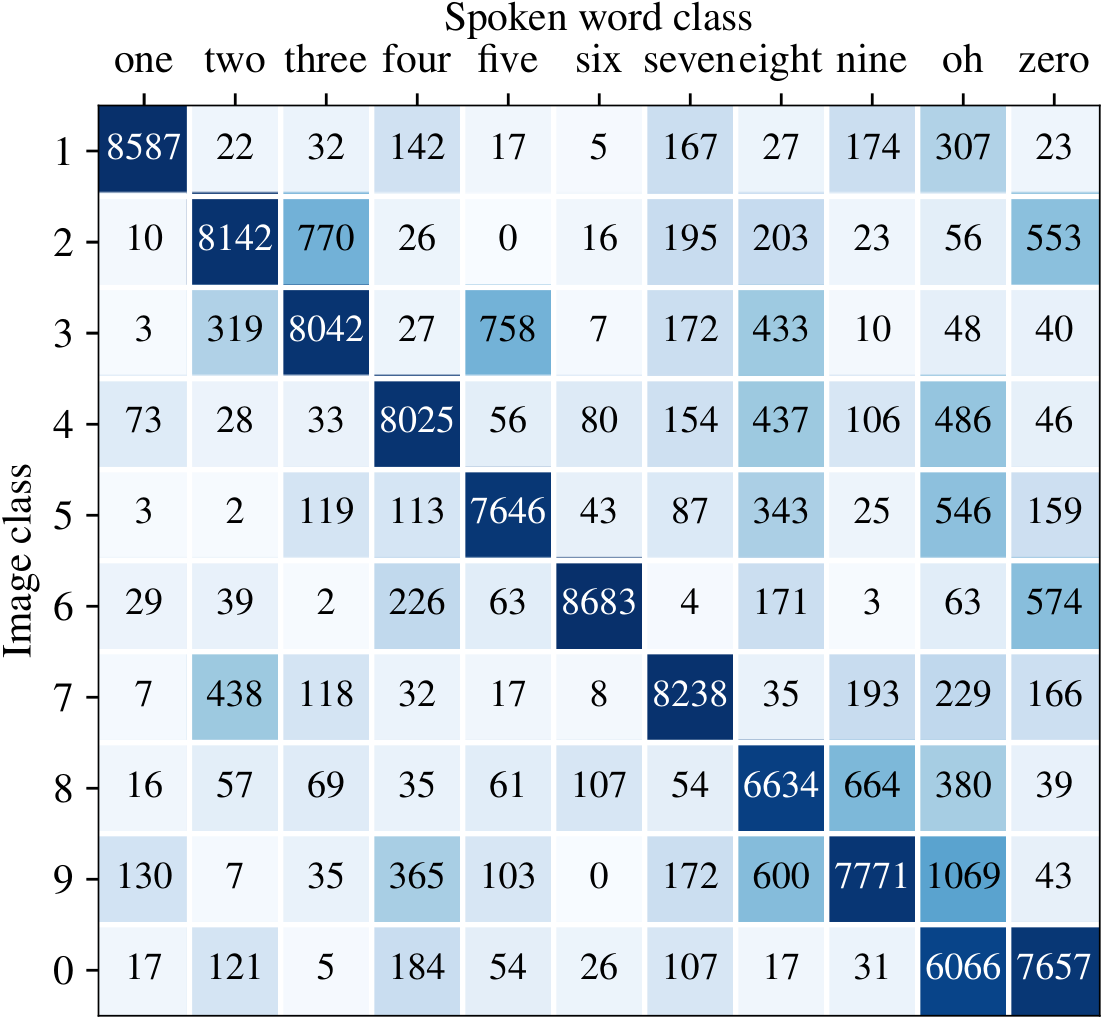}
	\vspace*{-7.5pt}
	\caption{Confusion matrix counts for the \ac{MTriplet} from Table~\ref{tbl:multimodal}.}
	\label{fig:confusions}
	\vspace*{-18pt}
\end{figure}

Finally we are interested in the types of mistakes that the direct models make.
Figure~\ref{fig:confusions} shows the confusion matrix for the \ac{MTriplet} from Table~\ref{tbl:multimodal}.
We see that the most confusions are for the spoken word ``oh'' with the image $9$.
Qualitative analysis shows that this is because many $0$ and $9$ images are subjectively similar, rather than ``oh'' and ``nine'' being acoustically confusable.
%Further support that confusions are mostly due to image representation comparisons is that when the multimodal models are used for \textit{unimodal} few-shot classification, scores in image classification are generally worse than for speech classification.
The fact that confusions are mostly due to image comparisons is supported by using the multimodal models to do \textit{unimodal} few-shot classification: image classification are generally worse than speech classification (scores not shown~here).

%% file: Sections/Conclusion.tex
\section{Conclusion}

We proposed two novel direct multimodal few-shot learning models which outperformed existing indirect multimodal models by a substantial margin. 
An \ac{MTriplet} model performed best.
We showed that the improvements over indirect models are due to the combination of unsupervised and transfer learning, which results in more accurate embedding spaces and do not require two comparisons (i.e.\ no compounding of mistakes).
Future work will look into improving the pair mining process and the feasibility of using these models on more realistic datasets.

\vspace{2pt}
{\eightpt
\noindent \textbf{Acknowledgements.} This work is supported in part by the National Research Foundation of South Africa (grant no.\ 120409), CSIR and Saigen Scholarships for LN, and a Google Faculty Award for HK.}